# Finite-Time Analysis of Kernelised Contextual Bandits

**Michal Valko, Nathan Korda, Rémi Munos**
SequeL team
INRIA Lille - Nord Europe, France

**Ilias Flaounas, Nello Cristianini**
Intelligent Systems Laboratory
University of Bristol, UK

## Abstract

We tackle the problem of online reward maximisation over a large finite set of actions described by their contexts. We focus on the case when the number of actions is too big to sample all of them even once. However we assume that we have access to the similarities between actions' contexts and that the expected reward is an *arbitrary* linear function of the contexts' images in the related reproducing kernel Hilbert space (RKHS). We propose KernelUCB, a kernelised UCB algorithm, and give a cumulative regret bound through a frequentist analysis. For contextual bandits, the related algorithm GP-UCB turns out to be a special case of our algorithm, and our finite-time analysis improves the regret bound of GP-UCB for the agnostic case, both in the terms of the kernel-dependent quantity and the RKHS norm of the reward function. Moreover, for the linear kernel, our regret bound matches the lower bound for contextual linear bandits.

## 1 Introduction

There are many situations in which an environment repeatedly provides an agent with a very large number of actions together with some contextual information (Cesa-Bianchi & Lugosi, 2006). These actions yield rewards when chosen and the agent wants to continually choose actions that yield high expected reward while not having enough time to explore them all. Thus it is natural to learn a relationship between the context provided for each action and the expected reward it produces. Kernel methods (Shawe-Taylor & Cristianini, 2004) provide a way to extract from observations possibly non-linear relationships between the contexts and the rewards while only using similarity information between contexts. In many applications similarity information is cheaply computable. In some situations the contexts are not even available and instead only similarities are given (Chen, Garcia, Gupta, Rahimi, & Cazzanti, 2009).

A typical example (Li, Chu, Langford, & Schapire, 2010), is the case of online advertisement in which one needs to continually show the most relevant ads to users viewing a website; since there is a simple binary reward of 1 for a click on the ad shown and 0 otherwise it is always costly to show ads that have only a small chance of being clicked on. Another example is a recommender system for relevant content from a large number of available news feeds (Steinberger, Pouliquen, & Van der Goot, 2009); here it is assumed that we can assess the relevance of the content of a feed based on information such as the anchor text of the feed link without having to get and process the actual feed content, which is a costly operation.

Our modelling assumption is that the expected reward obtained from choosing an action is a function of the features associated with that action. In the advertisement example the features are built from webpage content and user attributes. In the news feeds, the features come from easily retrievable information such as URLs, feed titles, or anchor text. We refer to the features as *contexts* and to the resulting problems of maximising cumulative reward as *contextual bandit problems*. One aspect that makes this setting different from related settings is the possibly changing decision set.

Previous approaches (Li et al., 2010; Chu, Li, Reyzin, & Schapire, 2011; Auer, 2002) to contextual bandit problems have often assumed that the functional relationship between the features and the expected rewards is linear. However the availability of similarity information gives us the opportunity to search for a linear relationship in a reproducing kernel Hilbert space (RKHS) defined by these similarities to discover a non-linear relationship between the context and the reward. Recently, Srinivas, Krause, Kakade, and Seeger

(2010) proposed the GP-UCB algorithm that optimises a function $\theta^*$ sampled from a Gaussian Process (GP) prior. In this paper we take an agnostic approach (Table 1) and provide the KernelUCB algorithm which comes directly from kernelising contextual linear bandits. KernelUCB is a kernel-based upper confidence bound algorithm which, given the similarity between two data points, uses the dualisation of regularised linear regression in the RKHS to find upper confidence bounds on the expected rewards of each action, and then chooses an action with the highest upper confidence bound. When the kernel is just the dot product between feature vectors KernelUCB is identical to LinUCB (Li et al., 2010), i.e., KernelUCB is a non-linear extension of LinUCB.

Our main contribution is a theoretical analysis of this approach. While kernelisation of linear bandits is straighforward, the analysis has to deal with an RKHS with potentially infinite dimension. We provide a data-dependent performance bound based on a notion of the effective dimension $\tilde{d}$. This quantity roughly measures the number of directions in the RKHS along which the data mostly lies. We are able to provide a cumulative regret bound that scales as $\tilde{O}(\sqrt{T\tilde{d}})$, where $T$ is the time and $\tilde{O}$ hides log factors. When the kernel is just the dot product between contexts, $\tilde{d}$ is upper bounded by the dimension of the contexts, and we recover the regret bounds for LinUCB for contextual linear bandits as a special case. The GP-UCB algorithm is also a special case of KernelUCB when the regulariser is set to the model noise, and we make (Section 4.1) a clear comparison with the *agnostic* analysis of GP-UCB (Srinivas et al., 2010), i.e. when their reward function $\theta^*$ is not sampled from a GP. For this agnostic case, Srinivas et al. (2010) obtain a cumulative regret bound $\tilde{O}(I(y_T; \theta^*)\sqrt{T})$ where $I(y_T; \theta^*)$ is the information gain between $\theta^*$ and the observed samples $y_T$. We show that $I(y_T; \theta^*)$ is $\Omega(\tilde{d})$ and since our bound only scales with $\sqrt{\tilde{d}}$, our analysis matches the lowerbound for the linear case, unlike the agnostic analysis of GP-UCB. Furthermore, due to the link between $\tilde{d}$ and $I(y_T; \theta^*)$ we can provide the *data-independent* worst case upperbounds for the popular kernels (such as RBF) by plugging the upperbounds $I(y_T; \theta^*)$ derived by Srinivas et al. (2010) into our improved analysis. Our analysis also gives us a guideline on how to set the regularisation parameter.

Section 2 presents the basic linear contextual bandit model and related work. In Section 3 we derive the KernelUCB algorithm by directly kernelising contextual linear bandits. In Section 4 we analyse KernelUCB, provide an upper bound on the cumulative regret and describe the tradeoff between the regularization and the RKHS norm of the reward function.

|  | **Bayesian** | **Frequentist** |
|---|---|---|
| **regression** | GP-Regression | Kernel Ridge Regression |
| **bandits** | GP-UCB | **KernelUCB** this paper |

Table 1: Bayesian and frequentist approaches to kernelized regression and contextual bandits

## 2 Background

### 2.1 Basic Model

We describe the basic settings and goals of linear contextual bandit problems. At each time $t$, for each action $a \in \mathcal{A} := \{1, \dots, N\}$, there is an associated context vector $x_{a,t} \in \mathbb{R}^d$. If action $a$ is chosen at time $t$ we have $a_t = a$ and receive a reward $r_{a,t}$ drawn from a distribution $\nu_{a,x_{a,t}}$. An algorithm $\pi$ is a method for choosing an action at time $t$ given the history i.e., the previously observed contexts, actions and rewards, and the current context:

$$H_{t-1} := \left( \{x_{a,j}\}_{a \in \mathcal{A}}, a_j, r_{a_j,j} \right)_{j<t} \cup \{x_{a,t}\}_{a \in \mathcal{A}},$$

$$\pi : H_{t-1} \mapsto \pi_t \in \mathcal{P}(\mathcal{A}),$$

where $\mathcal{P}(\mathcal{A})$ denotes the set of probability distributions over $\mathcal{A}$. For simplicity, we define $x_t := x_{a_t,t}$ and $r_t := r_{a_t,t}$ to be the context and the reward at the time $t$.

In the case of classical bandits, the reward distributions $\nu_{a,x_{a,t}}$ are independent of the context vectors, $x_{a,t}$. In this case we define the optimal action as $a^* := \arg\max_{a \in \mathcal{A}} \{\mathbb{E}(r_a)\}$ and define the regret of an algorithm at time $T$ to be:

$$R(T) := \sum_{t=1}^{T} r_{a^*,t} - r_t.$$

For linear contextual bandits we assume a linear relationship between contexts and mean rewards,

$$\mathbb{E}[r_{a,t} \mid x_{a,t}] = x_{a,t}^\intercal \theta^*,$$

for some fixed but unknown vector $\theta^* \in \mathbb{R}^d$. Note, that $\theta^*$ is the same for all actions and thus this problem is also called a fixed design setting (Bubeck & Cesa-Bianchi, 2012). In some (noncontextual) linear bandit settings (Dani, Hayes, & Kakade, 2008), the contexts do not change and $x_{a,t} = x_a$.

In this paper we consider the case when the contexts, and subsequently the optimal action, *can change* over time. Thus we have $a_t^* := \arg\max_{a \in \mathcal{A}} \{\mathbb{E}(r_{a,t} \mid x_{a,t})\}$

and the (contextual) regret of an algorithm at time $T$ becomes:

$$R(T) := \sum_{t=1}^{T} r_{a_t^*, t} - r_t. \qquad (1)$$

The aim in both of these situations is to find an algorithm which minimises the regret at time $T$.

## 2.2 UCB Algorithms

Upper confidence bound (UCB) algorithms (Lai & Robbins, 1985) provide a simple but efficient heuristic approach to bandit problems. The central idea is to maintain for each action, $a$, an estimate of the mean reward $\hat{\mu}_{a,t}$ and a confidence interval around that mean with width $\hat{\sigma}_{a,t}$. At each time $t$ the algorithm then chooses the action with the highest upper confidence bound $\hat{\mu}_t + \hat{\sigma}_{a,t}$; thus an action $a$ is selected if either it has a high estimated mean, or if there is much uncertainty about the action so that the width $\hat{\sigma}_{a,t}$ is large.

For classical bandits (with no contextual information) it is possible to obtain finite time analyses of such algorithms along the following lines: Construct the widths $\hat{\sigma}_{a,t}$ so that they are large when $a$ has not been played often but small when it has been played a large number of times already, for example by relating them to the standard deviations of the estimates $\hat{\mu}_{a,t}$. Assume that a suboptimal action, $a$, has been played a large number of times. Then through tools such as the Azuma-Hoeffding inequality one can expect to obtain high probability bounds on the events that $\hat{\mu}_{a,t}$ is close to $\mu_a$. From the construction of the widths it follows that $\hat{\mu}_{a,t} + \hat{\sigma}_{a,t}$ will also be close to $\mu_a$. In this way as soon as a sub-optimal action $a$ has been played enough times so that $\hat{\mu}_{a,t} + \hat{\sigma}_{a,t} < \hat{\mu}_{a^*}$, the probability this action will be played again becomes very small. Such analyses typically conclude that UCB algorithms are close to optimal, and they motivate the choice of widths relating to the standard deviations of the estimates $\hat{\mu}_{a,t}$.

## 2.3 UCBs for Linear Contextual Bandits

Since we assume that there is a functional relationship between the expected rewards of an action and the feature vectors observed, constructing the estimates $\hat{\mu}_{a,t}$ and the widths $\hat{\sigma}_{a,t}$ can be approached by regression. In particular, when we assume a linear model we can use regularised least squares regression to estimate the mean rewards:

$$\hat{\mu}_{a,t} := x_{a,t}^\intercal \hat{\theta}_t$$

where $\hat{\theta}_t := C_t^{-1} X_t^\intercal y_t$, $y_t := \{r_{a_1,t}, \dots, r_{a_t,t}\}^\intercal$, $X_t := \{x_{a,1}, \dots, x_{a_t,t}\}^\intercal$, and $C_t := X_t^\intercal X_t + \gamma I_d$ for some

$\gamma > 0$. Appropriate widths for the confidence intervals can be described in terms of the Mahalanobis distance of $x_{a,t}$ from the centre of mass of $X_t$:

$$\hat{\sigma}_{a,t} = \sqrt{x_{a,t}^\intercal C_t^{-1} x_{a,t}}$$

These widths relate to variance in the data: For instance in the case of standard normal noise (i.e. when the rewards satisfy $r_{a,t} = x_{a,t}^T \theta^* + \varepsilon_{a,t}$, where all $\varepsilon_{a,t} \sim \mathcal{N}(0,1)$), $\hat{\sigma}_{a,t}^2$ is exactly the variance of $\hat{\mu}_{a,t}$. Even when no assumption is made on the noise, this Mahalanobis distance has the property of being small when $x_{a,t}$ is close to the center of mass of data $X_t$, and large otherwise. Consequently a generic UCB type algorithm based on the estimators $\hat{\mu}_{a,t}$ and $\hat{\sigma}_{a,t}^2$ chooses an action $a_t$ at time $t$ such that:

$$a_t = \arg\max_{a \in A} \left( x_{a,t}^\intercal C_t^{-1} X_t^\intercal y_t + \eta \sqrt{x_{a,t}^\intercal C_t^{-1} x_{a,t}} \right),$$

where $\eta = \eta(t)$ is some (possibly time dependent) deterministic parameter of the algorithm which we call the *exploration* parameter.

Based on these ideas, Li et al. (2010) propose LinUCB which treats $\eta(t) = \eta$ as a constant that needs to be optimised. While this algorithm is simple to understand and implement in practice, no optimal theoretical regret analysis exists in the literature for LinUCB. Instead Chu et al. (Chu et al., 2011) give a theoretical analysis of a related algorithm, SupLinUCB, and achieve with probability $1 - \delta$ a regret bound of:

$$O \left( \sqrt{T d \ln^3(NT \ln(T)/\delta)} \right).$$

## 2.4 Related Work

The most related work to our setting is LinUCB (Li et al., 2010) and SupLinUCB (Chu et al., 2011), which were inspired by SupLinRel (Auer, 2002), an early algorithm for linear contextual bandits. Instead of using regularised linear regression SupLinRel uses eigendecomposition to make a pseudo-inverse of the covariance matrix. A discussion of practical advantages of SupLinUCB over SupLinRel can be found in (Li et al., 2010). SupLinRel achieves a regret bound $O((Td \ln^{3/2}(2NT \ln(T)/\delta))^{1/2})$.

Interestingly, one can derive an instantiation of KernelUCB in the Bayesian setting. This is the case of GP-UCB (Srinivas et al., 2010) a special case of KernelUCB, which assumes that the reward function is drawn from a GP prior. The conceptual difference between the KernelUCB and GP-UCB is similar to the difference between kernel regression and GP-regression (Table 1). Nevertheless, Srinivas et al. (2010) also provide a frequentist analysis of GP-UCB, which we compare to in Section 4.1. Krause and Ong (2011) later propose CGP-UCB, an extension of GP-UCB for the setting when each action has its own intrinsic features, as well as features associated to its changing environment. It therefore uses possibly different kernels for the action and the context spaces.

Slivkins (2009) takes advantage of similarity information between contexts, where he builds on previous work (Kleinberg, Slivkins, & Upfal, 2008; Lu, Pál, & Pál, 2010) that assume only a metric space structure on the context and action spaces. The setting in (Slivkins, 2009) is different from ours: they assume a Lipschitz property in a similarity space, which is a weaker condition than in our setting, but as a consequence their bound depends more heavily on the relevant dimensions (the covering dimensions of the context and action spaces appears in the exponent of $T$ whereas our effective dimension appears as a multiplicative factor only).

Another well known related family are the Confidence-Ball algorithms (Abbasi-Yadkori, Pal, & Szepesvari, 2011; Dani et al., 2008; Rusmevichientong & Tsitsiklis, 2010). These solve the linear bandit problem in which the action space is the context space and there is a reward linear in contexts. When we fix the contexts in our own setting we recover the linear bandit model for a finite action set and in that sense our setting is more general. For continuous action space linear bandits the attainable lower bound for the regret is $\Omega(d\sqrt{T})$ (Dani et al., 2008), whereas for finite action space linear contextual bandits the attainable lower bound on regret is $\Omega(\sqrt{dT})$ (Chu et al., 2011).

A set of algorithms based on EXP4 (Auer, Cesa-Bianchi, Freund, & Schapire, 2003) such as EXP4.P (Beygelzimer, Langford, Li, Reyzin, & Schapire, 2010) or Policy Elimination (Dudik, Hsu, Kale, Karampatziakis, Langford, Reyzin, & Zhang, 2011) can deal with the general case of an arbitrary set of hypotheses together with finite action sets. Their definition of regret is different from ours since they compare to the best fixed-parameter solution, whereas we compare to the best action with respect to the changing context. For a general discussion of the advantages of approaches directly taking advantage of structure in contextual bandit problems over the EXP4 family we refer to (Chu et al., 2011). Epoch-Greedy (Langford & Zhang, 2008), which also works in a setting more general than ours, achieves a better dependence on the size of the set of hypotheses but a worse dependence on time $T$. The VE algorithm (Beygelzimer et al., 2010) which is based on EXP4.P has a regret bound that scales as $O(\sqrt{Td\ln T})$ where $d$ is the VC dimension of the hypothesis class.

Other related work includes (Seldin, Auer, Laviolette, Shawe-Taylor, & Ortner, 2011) which studies a different setting with finite context spaces, showing a regret bound that depends on the mutual information between contexts and actions, and Gaussian process bandits (Grünewälder, Audibert, Opper, & Shawe-Taylor, 2010) and convex bandits (Cesa-Bianchi & Lugosi, 2006) study mostly continuous actions sets.

## 3 Kernelised UCB

In this section we show how to derive KernelUCB by directly kernelising the LinUCB algorithm. In contrast GP-UCB is motivated from experimental design. The derivation is straightforward and we provide it for convenience and to introduce the notation which is used in the analysis. Our derivation is the combination of the kernel trick (Shawe-Taylor & Cristianini, 2004) and the kernelised version of the Mahalanobis (Haasdonk & Pekalska, 2010).

Kernel methods assume that there exists a mapping $\phi : \mathbb{R}^d \to \mathcal{H}$ that maps the data to a (possibly infinite dimensional) Hilbert space in which a linear relationship can be observed. We call $\mathbb{R}^d$ the *primal space* and $\mathcal{H}$ the associated *reproducing kernel Hilbert space* (RKHS). We use matrix notation to denote the inner product of two elements $h, h' \in \mathcal{H}$, i.e. $h^\intercal h' := \langle h, h' \rangle_{\mathcal{H}}$ and $\|h\| = \sqrt{\langle h, h \rangle_{\mathcal{H}}}$ to denote the RKHS norm. From the mapping $\phi$ we have the *kernel function*, defined by:

$$k(x, x') := \phi(x)^\intercal \phi(x'), \ \forall x, x' \in \mathbb{R}^d,$$

and the *kernel matrix* of a data set $\{x_1, \ldots, x_t\} \subset \mathbb{R}^d$ given by $K_t := \{k(x_i, x_j)\}_{i,j \leq t}$. For our non-linear contextual bandit model we assume the existence of a $\phi$ for which there exists a $\theta^* \in \mathcal{H}$ such that:

$$\mathbb{E}(r_{a,t} \mid x_{a,t}) = \phi(x_{a,t})^\intercal \theta^*.$$

Taking $a_t^* := \arg\max_{a \in \mathcal{A}}\{\phi(x_{a,t})^\intercal \theta^*\}$ we can define the regret as before in (1). Note that when $\phi \equiv \text{Id}$, we recover the linear bandit case.

To obtain the upper confidence bounds we derive prediction and width estimators for the expected rewards. LinUCB uses estimators built from ridge regression in the primal. Since we assume that our model is linear in the RKHS we show how to build estimators from ridge regression in $\mathcal{H}$. By deriving equivalent dual forms which involve only entries of the kernel matrix we avoid working directly in the possibly infinite dimensional RKHS.

First we take the prediction estimator to be of the form $\hat{\mu}_{a,t+1} = \phi(x_{a,t+1})^\intercal \theta_t$ where $\theta_t$ is the minimiser of the regularised least squares loss function:

$$\mathcal{L}(\theta) = \gamma \|\theta\|^2 + \sum_{i=1}^{t-1} (r_i - \phi(x_i)^\intercal \theta)^2. \qquad (2)$$

We derive a representation of this estimator involving only kernels between context vectors. We denote $\Phi_t = [\phi(x_1)^\intercal, \ldots, \phi(x_{t-1})^\intercal]^\intercal$. Note that the solution of the minimisation problem $\theta_t := \min_{\theta \in \mathcal{H}} \mathcal{L}(\theta)$ satisfies:

$$(\Phi_t^\intercal \Phi_t + \gamma I)\theta_t = \Phi_t^\intercal y_t.$$

Rearranging this equation we obtain:

$$\theta_t = \Phi_t^\intercal \alpha_t \qquad (3)$$

where $\alpha_t = \gamma^{-1}(y_t - \Phi_t \theta_t) = \gamma^{-1}(y_t - \Phi_t \Phi_t^\intercal \alpha_t)$, which implies that $\alpha_t = (K_t + \gamma I)^{-1} y_t$. Finally, denoting $k_{x,t} := \Phi_t \phi(x) = [k(x, x_1), \ldots, k(x, x_{t-1})]^\intercal$ we get:

$$\hat{\mu}_{a,t} = k_{x_{a,t},t}^\intercal (K_t + \gamma I)^{-1} y_t. \qquad (4)$$

While the computation of $\theta_t$ using (3) would require evaluating $\phi(x_i)$ for every data point $x_i$, the dualised representation of the prediction (4) allows the computation of $\hat{\mu}_{a,t}(x)$ only from objects in the kernel matrix.

Next we construct the widths of the confidence intervals around the prediction. As for linear bandits we find appropriate widths in terms of the Mahalanobis distance of $\phi(x_{a,t})$ from the matrix $\Phi_t$:

$$\hat{\sigma}_{a,t} := \sqrt{\phi(x_{a,t})^\intercal (\Phi_t^\intercal \Phi_t + \gamma I)^{-1} \phi(x_{a,t})}. \qquad (5)$$

Once again we motivate this choice of width by noting that it is exactly the variance of the prediction estimator when the noise in the dualised data is standard normal. In order to compute these widths we derive a dualised representation of (5). Our derivation is similar to the kernelisation of the Mahalanobis distance for centered data in (Haasdonk & Pekalska, 2010): Since the matrices $(\Phi_t^\intercal \Phi_t + \gamma I)$ and $(\Phi_t \Phi_t^\intercal + \gamma I)$ are regularised they are strictly positive definite, and therefore:

$$(\Phi_t^\intercal \Phi_t + \gamma I)\Phi_t^\intercal = \Phi_t^\intercal (\Phi_t \Phi_t^\intercal + \gamma I)$$

$$\Phi_t^\intercal (\Phi_t \Phi_t^\intercal + \gamma I)^{-1} = (\Phi_t^\intercal \Phi_t + \gamma I)^{-1} \Phi_t^\intercal.$$

Now we can extract the Mahalanobis distance from the last equation

$$(\Phi_t^\intercal \Phi_t + \gamma I)\phi(x) = (\Phi_t^\intercal k_{x,t} + \gamma \phi(x))$$

from which we deduce that

$$\phi(x) = \Phi_t^\intercal (\Phi_t \Phi_t^\intercal + \gamma I)^{-1} k_{x,t} + \gamma (\Phi_t^\intercal \Phi_t + \gamma I)^{-1} \phi(x)$$

and express $\phi(x)^\intercal \phi(x)$ as

$$k_{x,t}^\intercal (\Phi_t \Phi_t^\intercal + \gamma I)^{-1} k_{x,t} + \gamma \phi(x)^\intercal (\Phi_t^\intercal \Phi_t + \gamma I)^{-1} \phi(x).$$

Rearranging we get an expression for the width involving only inner products:

$$\hat{\sigma}_{a,t} := \gamma^{-1/2} \sqrt{k(x_{a,t}, x_{a,t}) - k_{x_{a,t},t}^\intercal (K_t + \gamma I)^{-1} k_{x_{a,t},t}}.$$

---

**Algorithm 1** KernelUCB with online updates

**Input and initialisation:**
    $N$ the number of actions, $T$ the number of pulls,
    $\gamma$, $\eta$ regularization and exploration parameters
    $k(\cdot, \cdot)$ kernel function
    $u_0 \leftarrow [1, 0, ..., 0]^\intercal$ (at start first action is pulled)
    $y_0 \leftarrow \emptyset$
**Run:**
**for** $t = 1$ **to** $T$ **do**
    Choose $a \leftarrow \arg\max u_{t-1}$ and get reward $r_{t-1}$
    Update $y_t \leftarrow [r_1, \ldots, r_{t-1}]^\intercal$
    **if** $t = 1$ **then**
        $K_t^{-1} \leftarrow 1/k_{x_t,x_t} + \gamma$
    **else** {online update of the kernel matrix inverse}
        $b \leftarrow (k_{x_1}, k_{x_2}, \ldots, k_{x_{t-1}})^\intercal$
        $K_{22} \leftarrow (k_{x_a, x_a} + \gamma - b^\intercal K_{t-1}^{-1} b)^{-1}$
        $K_{11} \leftarrow K_{t-1}^{-1} + K_{22} K_{t-1}^{-1} b b^\intercal K_{t-1}^{-1}$
        $K_{12} \leftarrow -K_{22} K_{t-1}^{-1} b$
        $K_{21} \leftarrow -K_{22} b^\intercal K_{t-1}^{-1}$
        $K_t^{-1} \leftarrow [K_{11}, K_{12}; K_{21}, K_{22}]$
    **end if**
    **for** $a = 1$ **to** $N$ **do**
        $\sigma_{a,t} \leftarrow \sqrt{k(x_{a,t}, x_{a,t}) - k_{x,t}^\intercal K_t^{-1} k_{x,t}}$
        $u_{a,t} \leftarrow \left(k_{x,t}^\intercal K_t^{-1} y_t + \frac{\eta}{\gamma^{1/2}} \sigma_{a,t}\right)$
    **end for**
**end for**

---

As for LinUCB, KernelUCB chooses the action $a_t$ at time $t$ which satisfies

$$a_t := \arg\max_{a \in A} (k_{x_{a,t},t}^\intercal (K_t + \gamma I_t)^{-1} y_t +$$
$$+ \frac{\eta}{\gamma^{1/2}} \sqrt{k(x_{a,t}, x_{a,t}) - k_{x_{a,t},t}^\intercal (K_t + \gamma I)^{-1} k_{x_{a,t},t}}),$$

where $\eta$ is a (possibly time dependent) exploration parameter of the algorithm. Considering $a_t$ and $\hat{\sigma}_{a,t}$ we see that GP-UCB is a special case of KernelUCB where the regularization constant is set to the model noise.

The selection of an appropriate kernel function is problem dependent (Shawe-Taylor & Cristianini, 2004). The linear kernel corresponds to $\phi \equiv \text{Id}$ and leads to the dual representation of the LinUCB algorithm in the primal. A non-linear kernel function creates a kernelised UCB algorithm for a non-linear bandit. Typical examples of non-linear kernel functions include: the radial basis function where $k(x_i, x_j) = \exp(-||x_i - x_j||^2/2\sigma^2)$, for $\sigma > 0$ and the polynomial kernel $k(x_i, x_j) = (x_i^\intercal x_j + 1)^p$. The pseudocode of KernelUCB is displayed in Algorithm 1 and uses the inversion update of $K_t$ through the properties of the Schur complement (Zhang, 2005).

**Algorithm 2** SupKernelUCB

**Input and initialisation:**
    $T$ number of arm pulls, $S$ number of sets
  $\Psi_1^{(s)} \leftarrow \emptyset$ for all $s \in [T]$
  **for** $t = 1$ **to** $T$ **do**
    $s \leftarrow 1$ and $\hat{A}_1 \leftarrow [N]$
    **repeat**
      $\left(\hat{\mu}_{t,a}^{(s)}, \hat{\sigma}_{t,a}^{(s)}\right) \leftarrow$ BaseKernelUCB with $\Psi_t^{(s)}$ for all $a \in \hat{A}_{(s)}$
      **if** $\eta \hat{\sigma}_{t,a}^{(s)} \leq 1/\sqrt{T}$ for all $a \in \hat{A}_{(s)}$ **then**
        Choose $a_t = \arg\max_{a \in \hat{A}_{(s)}} \left(\hat{\mu}_{t,a}^{(s)} + \eta \hat{\sigma}_{t,a}^{(s)}\right)$
        Keep the sets $\Psi_{t+1}^{(s')} = \Psi_t^{(s')}$ for all $s' \in [S]$
      **else**
        **if** $\eta \hat{\sigma}_{t,a}^{(s)} \leq 2^{-s}$ for all $a \in \hat{A}_{(s)}$ **then**
          $\hat{A}_{(s+1)} \leftarrow \{a \in \hat{A}_{(s)} | \hat{\mu}_{t,a}^{(s)} + \eta \hat{\sigma}_{t,a}^{(s)} \geq \max_{a' \in \hat{A}_{(s)}} \hat{\mu}_{t,a'}^{(s)} + \eta \hat{\sigma}_{t,a'}^{(s)} - 2^{1-s}\}$
          $s \leftarrow s + 1$
        **else**
          Choose $a_t \in \hat{A}_{(s)}$ such that $\eta \hat{\sigma}_{t,a}^{(s)} > 2^{-s}$.
          Update the index sets at all levels $\Psi_{t+1}^{(s')} =$
          $\begin{cases} \Psi_t^{(s')} \cup \{t\} & \text{if } s = s \\ \Psi_t^{(s')} & \text{otherwise.} \end{cases}$
        **end if**
      **end if**
    **until** an action $a_t$ is found
  **end for**

## 4  Analysis

In this section we provide an upper bound on the cumulative regret defined in Section 2 for KernelUCB. As for LinUCB, the predictors for KernelUCB, $\hat{\mu}_{a,t}$, are sums of *dependent* random variables. Consequently, we are unable to directly apply the Azuma-Hoeffding inequality to gain control over the error in the predictors. To get around this problem we use the construction of Auer (2002) and introduce the related algorithm SupKernelUCB, the appropriate modification of KernelUCB. SupKernelUCB (Algorithm 2) constructs special, mutually exclusive subsets $\{\Psi_t^{(s)}\}_s$ of the elapsed time. On each of these sets it builds predictors, $\hat{\mu}_{t,a}^{(s)}$, and widths, $\hat{\sigma}_{a,t}^{(s)}$, in the same way that KernelUCB does, using the BaseKernelUCB (Algorithm 3) subroutine. In the pseudocodes and below, $[n]$ denotes the set $\{1, \ldots, n\}$. At the beginning of the algorithm all the subsets $\{\Psi_t^{(s)}\}_s$ are initialised to the empty set, and at each time $t \geq 1$ the value $t$ is included in at most one $\{\Psi_{t+1}^{(s)}\}_s$ in such a way that the event $\{t \in \Psi_{t+1}^{(s)}\}$ is independent of the rewards observed at times in $\Psi_t^{(s)}$. In this way the Azuma-

**Algorithm 3** BaseKernelUCB

**Input and initialisation:**
  $\Psi_t \subseteq \{1, 2, \ldots, t-1\}$
  $k(\cdot, \cdot)$ kernel function, $\gamma$ regularization parameter
  $K \leftarrow [k(x_i, x_j)]_{i,j \in \Psi_t} + \gamma I$
  $y \leftarrow [r_\tau]_{\tau \in \Psi_t}$
  **for** $a \in [N]$ **do**
    $\hat{\mu}_{t,a} \leftarrow k_{x_{a,t}}^\mathsf{T} K^{-1} y_t$
    $\hat{\sigma}_{t,a} \leftarrow \gamma^{-1/2} \sqrt{k(x_{a_t}, x_{a_t}) - x_{a_t}^\mathsf{T} K^{-1} x_{a_t}}$
  **end for**

Hoeffding inequality can be applied on each subset $\Psi_t^{(s)}$ to get a regret bound.

If we directly applied known regret bounds (Auer, 2002; Chu et al., 2011) for linear contextual bandits to our setting, we would obtain a bound in terms of the dimension of the RKHS, which is possibly infinite. We avoid this problem through a careful consideration of the eigenvalues of the covariance matrix and the choice of the regularisation constant and give a bound in terms of a data dependent quantity $\tilde{d}$ which we call the *effective dimension*: Let $(\lambda_{i,t})_{i \geq 1}$ denote the eigenvalues of $C_t^\gamma = \Phi_t^\mathsf{T} \Phi_t + \gamma I$ in decreasing order and define:

$$\tilde{d} := \min\{j : j\gamma \ln T \geq \Lambda_{T,j}\} \text{ where } \Lambda_{T,j} := \sum_{i > j} \lambda_{i,T} - \gamma.$$

**Theorem 1** *Assume that $\|\phi(x_{a,t})\| \leq 1$ and $|r_{a,t}| \in [0,1]$ for all $a \in A$ and $t \geq 1$, and set $\eta = \sqrt{2\ln 2TN/\delta}$. Then with probability $1 - \delta$, SupKernelUCB satisfies:*

$$R(T) \leq \left[2 + 2\left(1 + \sqrt{\frac{\gamma}{2\ln(2TN(1+\ln T)/\delta)}}\right)\|\theta^*\| + \right.$$
$$+ 8\sqrt{\left(12 + \frac{15}{\gamma}\right)\max\left\{\ln\left(\frac{T}{\tilde{d}\gamma} + 1\right), \ln T\right\}^3} \times$$
$$\left. \times \sqrt{\left(2\ln\frac{2TN(1+\ln T)}{\delta}\right)}\right]\sqrt{\tilde{d}T}$$

**Remark 1** *We call $\tilde{d}$ the effective dimension because it gives a proxy for the number of principle directions over which the projection of the data in the RKHS is spread. If the data all fall within a subspace of $\mathcal{H}$ of dimension $d'$, then $\Lambda_{T,d'} = 0$ and $\tilde{d} \leq d'$. However more generally $\tilde{d}$ can be thought of as a measure of how quickly the eigenvalues of $\Phi_t^\mathsf{T} \Phi_t$ are decreasing. For example if the eigenvalues are only polynomially decreasing in $i$ (i.e. $\lambda_i \leq Ci^{-\alpha}$ for some $\alpha > 1$ and some constant $C > 0$) then $\tilde{d} \leq 1 + (C/(\gamma \ln T))^{1/\alpha}$.*

**Remark 2** *When $\Phi \equiv \mathrm{Id}$, $\tilde{d} \leq d$, the assumption that $\|\phi(x_{a,t})\| \leq 1$ becomes the assumption that the contexts* are normalised in the primal, and we recover exactly the result from (Chu et al., 2011) which matches the lower bound for this setting.

**Remark 3** *Theorem 1 suggests that if we know that $\|\theta^*\| \le L$, for some $L$, we should set $\gamma$ to be of the order of $L^{-1}$ so that we obtain $\tilde{O}(\sqrt{LdT})$ regret. If we do not have such knowledge, just setting $\gamma$ to a constant (e.g., found by a cross-validation) will incur $\tilde{O}(\|\theta^*\|\sqrt{\bar{d}T})$ regret.*

The proof of this theorem follows the scheme of the proof of Theorem 1 in (Chu et al., 2011). The first step is to prove a high probability bound on the error in the predictors $\hat{\mu}_{a,t}^{(s)}$, and to do this we use a classical concentration result, the Azuma-Hoeffding inequality. Our result here generalises Lemma 1 of Chu et al. (2011) to 1) linear products in RKHS, 2) regularisation $\gamma$, and 3) no assumption that $\|\theta^*\| \le 1$. Also the trade-off between $\gamma$ and $\|\theta^*\|$ becomes evident. For ease of notation, in the below we drop the superscript $(s)$ whenever it is superfluous.

**Lemma 1** *Suppose that the conditions of Theorem 1 hold, and that the input index set $\Psi_t^{(s)}$ for BaseKernelUCB is constructed so that for fixed contexts $x_{a_\tau,\tau}$, $\tau \in \Psi_t^{(s)}$, the rewards $r_{a_\tau,\tau}$ are independent random variables. Then with probability at least $1 - 2Ne^{-\eta^2/2}$ we have for all $a \in A$:*

$$|\hat{\mu}_{a,t}^{(s)} - \phi(x_{a,t})^\top \theta^*| \le (\eta(1 + \|\theta^*\|) + \gamma^{1/2}\|\theta^*\|)\hat{\sigma}_{a,t}^{(s)}.$$

**Proof.** We begin by noting that:

$$\hat{\mu}_{a,t} - \phi(x_{a,t})^\top \theta^* \quad (6)$$
$$= \phi(x_{a,t})^\top (C_t^\gamma)^{-1} \Phi_t^\top y_t - \phi(x_{a,t})^\top (C_t^\gamma)^{-1} (\Phi_t^\top \Phi_t + \gamma I) \theta^*$$
$$= \phi(x_{a,t})^\top (C_t^\gamma)^{-1} \Phi_t^\top (y_t - \Phi_t \theta^*) - \gamma \phi(x_{a,t})^\top (C_t^\gamma)^{-1} \theta^*$$

Now by construction of the set $\Psi_t$ we know that $(y_t - \Phi_t \theta^*) \mid \Phi_t, x_{a,t}$ is a vector of zero mean independent random variables. Hence we can apply the Azuma-Hoeffding inequality to obtain that:

$$\mathbb{P}(|\phi(x_{a,t})^\top (C_t^\gamma)^{-1} \Phi_t^\top (y_t - \Phi_t \theta^*)| \quad (7)$$
$$> (1 + \|\theta^*\|) \eta \hat{\sigma}_{a,t}) \le 2e^{-\eta^2/2},$$

since $|r_\tau - \phi(x_\tau)^\top \theta^*| \le 1 + \|\theta^*\|$ for any $\tau$ and

$$\hat{\sigma}_{a,t}^2 = \phi(x_{a,t})^\top (C_t^\gamma)^{-1} \phi(x_{a,t})$$
$$= \phi(x_{a,t})^\top (C_t^\gamma)^{-1} (\Phi_t^\top \Phi_t + \gamma I)(C_t^\gamma)^{-1} \phi(x_{a,t})$$
$$\ge \|\Phi_t (C_t^\gamma)^{-1} \phi(x_{a,t})\|^2.$$

Now by the Cauchy-Schwarz inequality we find that:

$$|\phi(x_{a,t})^\top (C_t^\gamma)^{-1} \theta^*| \le$$
$$\le \|\theta^*\| \sqrt{\phi(x_{a,t})^\top (C_t^\gamma)^{-1} \gamma^{-1} \gamma I (C_t^\gamma)^{-1} \phi(x_{a,t})}$$
$$\le \gamma^{-1/2} \|\theta^*\| \sqrt{\phi(x_{a,t})^\top (C_t^\gamma)^{-1} C_t^\gamma (C_t^\gamma)^{-1} \phi(x_{a,t})}$$
$$\le \gamma^{-1/2} \|\theta^*\| \hat{\sigma}_{a,t}. \quad (8)$$

The result follows by plugging (7) and (8) into (6). ■

The second step of the analysis bounds the widths $\hat{\sigma}_{a,t}^{(s)}$ in terms of the change in eigenvalues of the matrix $C_t^\gamma$. To do this we extend the argument in Lemma 11 by Auer (2002) to possibly infinite matrices. Let us define $\psi_{s,t} := |\Psi_t^{(s)}|$.

**Lemma 2** *The eigenvalues of $\Phi_t^\top \Phi_t$ do not depend on the choice of basis for $\mathcal{H}$. Moreover the representation of $\Phi_t^\top \Phi_t$ in any basis $\mathcal{B}$ created by extending a maximal linearly independent subset of $\{\phi(x_{a,\tau})\}_{\tau \in \Psi_t}$ has zeros everywhere outside its top-left ($\psi_t \times \psi_t$)-submatrix.*

**Proof.** Assume that $\phi = \phi_\mathcal{E}$ is described in terms of some basis $\mathcal{E}$ for $\mathcal{H}$. Let $\mathcal{B}$ be any basis for $\mathcal{H}$ extended from a maximal linearly independent subset of $\{\phi(x_{a,s})\}_{s \le t}$. If $Q_{\mathcal{B}\mathcal{E}}$ denotes the change of basis matrix from $\mathcal{B}$ to $\mathcal{E}$ then $\Phi_{\mathcal{E},t} = \Phi_{\mathcal{B},t} Q_{\mathcal{B}\mathcal{E}}$ and:

$$\Phi_{\mathcal{E},t}^\top \Phi_{\mathcal{E},t} = Q_{\mathcal{B}\mathcal{E}}^\top \Phi_{\mathcal{B},t}^\top \Phi_{\mathcal{B},t} Q_{\mathcal{B}\mathcal{E}},$$

where $\Phi_{\mathcal{B},t}$ and $\Phi_{\mathcal{E},t}$ denote the matrix $\Phi_t$ with respect to the bases $\mathcal{B}$ and $\mathcal{E}$. Moreover the $(i,j)$-th entry of $\Phi_{\mathcal{B},t}^\top \Phi_{\mathcal{B},t}$ is zero when $\max\{i,j\} > \psi_t$. Hence the eigenvalues are independent of the choice of basis and only the first $t$ of them can be non-zero. ■

**Lemma 3** *Suppose that $\Psi_{t+1}^{(s)} = \Psi_t^{(s)} \cup \{t\}$. Then the eigenvalues of $C_t^\gamma$ can be arranged so that $\lambda_{j,t-1} \le \lambda_{j,t}$ for each $j \ge 1$, and:*

$$\hat{\sigma}_{a,t}^{(s)} \le \sqrt{\left(4 + \frac{6}{\gamma}\right) \sum_{j=1}^{\psi_{s,t+1}} \frac{\lambda_{j,t} - \lambda_{j,t-1}}{\lambda_{j,t-1}}}.$$

*where $\lambda_{j,0} := \gamma$ for all $j$.*

**Proof.** Let $\mathcal{B}$ be a basis defined as in Lemma 2, let $C_{\mathcal{B},t} := \Phi_{\mathcal{B},t}^\top \Phi_{\mathcal{B},t}$, and let $\tilde{C}_t$ denote the top-left, $\psi_{s,t+1} \times \psi_{s,t+1}$-submatrix of $C_{\mathcal{B},t}$ and let $\tilde{\phi}(x_t)$ denote the first $t$ entries in the vector $\phi_{\mathcal{B}}(x_t)$. It follows from Lemma 2:

$$C_{\mathcal{B},t+1} + \gamma I = \left(\begin{array}{c|c} \tilde{C}_t & 0 \\ \hline 0 & 0 \end{array}\right) + \left(\begin{array}{c|c} \tilde{\phi}(x_t)\tilde{\phi}(x_t)^\top & 0 \\ \hline 0 & 0 \end{array}\right) + \gamma I$$

and we may apply the argument of the proof of Lemma 11 by Auer (2002) to the top left $\psi_{s,t+1} \times \psi_{s,t+1}$ blocks to obtain the result. Note that we only need to sum up to $\psi_{s,t+1}$ because $\lambda_j = \gamma$ for all $j > \psi_{s,t+1}$. ∎

The third step of the analysis uses the bound on the widths in Lemma 3 to bound their sum. Since our matrices $C_t^\gamma$ are possibly infinite we use the effective dimension of the data, $\tilde{d}$, to reduce the analysis to the finite dimensional case.

**Lemma 4** *Let $l_T = \max\{\ln(T/(\tilde{d}\gamma)+1), \ln T\}$. Then:*

$$\sum_{t \in \Psi_{T+1}^{(s)}} \hat{\sigma}_{a,t}^{(s)} \leq \sqrt{\left(10 + \frac{15}{\gamma}\right)\tilde{d}Tl_T} \text{ for all } s \in [S].$$

**Proof.** From Lemma 3 we know that the eigenvalues of $C_t^\gamma$ can be arranged so that $\lambda_{j,t-1} \leq \lambda_{j,t}$ for each $j \geq 1$. Once such an arrangement exists we can always rearrange the eigenvalues so that they are also decreasing in $j$, for each $t$. By Lemma 3 we have:

$$\sum_{t \in \Psi_{T+1}} \hat{\sigma}_{a,t} \leq \sum_{t=1}^{\psi_{T+1}} \sqrt{\sum_{j=1}^{\psi_{T+1}} \frac{\lambda_{j,t} - \lambda_{j,t-1}^{(s)}}{\lambda_{j,t-1}}}$$

$$\leq \sum_{t=1}^{\psi_{T+1}} \left[\sqrt{\sum_{1 \leq j \leq \tilde{d}} \frac{\lambda_{j,t} - \lambda_{j,t-1}}{\lambda_{j,t-1}}} + \sqrt{\sum_{j=\tilde{d}+1}^{\psi_{T+1}} \frac{\lambda_{j,t} - \lambda_{j,t-1}}{\lambda_{j,t-1}}}\right]$$

$$\leq \underbrace{\sum_{t=1}^{\psi_{T+1}} \sqrt{\sum_{1 \leq j \leq \tilde{d}} \frac{\lambda_{j,t} - \lambda_{j,t-1}}{\lambda_{j,t-1}}}}_{(A)} + \underbrace{\sqrt{\frac{\psi_{T+1}}{\gamma} \sum_{j \geq \tilde{d}+1} \lambda_{j,T}}}_{(B)},$$

where we have used the Cauchy-Schwarz inequality for the second inequality. Now by the definition of $\tilde{d}$, term (B) is bounded by $\sqrt{\tilde{d}\psi_{T+1}\ln T}$. Writing $\alpha_{i,t} = \lambda_{i,t} - \lambda_{i,t-1}$, term (A) becomes:

$$\sum_{t=1}^{\psi_{T+1}} \sqrt{\sum_{i=1}^{\tilde{d}} \frac{\alpha_{i,t}}{\sum_{s=1}^{t}\alpha_{i,s}+\gamma}}$$

where $\lambda_{i,0} = \gamma$ and $\sum_{i=1}^{\tilde{d}} \alpha_{i,t} \leq tr(C_t^\gamma) - tr(C_{t-1}^\gamma) = \|\phi(x_{a,t})\|^2 \leq 1$. We upper bound this object by solving an easier maximisation problem:

$$\max_{(\alpha)_{i,t},(\varepsilon)_{i,t}} \sum_{t=1}^{\psi_{T+1}} \sqrt{\sum_{i=1}^{\tilde{d}} \frac{\alpha_{i,t}}{\sum_{s=1}^{t}\varepsilon_{i,s}+\gamma}}$$

under the constraints $\sum_i \alpha_{i,t} = \sum_i \epsilon_{i,t} \leq 1$. Using the method of Lagrange multipliers and the Cauchy-Schwarz inequality[1] we can upper bound the solution to this maximisation problem by $\sqrt{\tilde{d}\psi_{T+1}\log(\psi_{T+1}/(\tilde{d}\gamma)+1)}$. We conclude by noting that $\psi_{T+1} \leq T$. ∎

---
[1] We do not include a detailed derivation due to the space constraints.

The fourth step is to bound the size of the sets $\Psi_{T+1}^{(s)}$. This is achieved by plugging our Lemma 4 into the proof of Lemma 16 in (Auer, 2002):

**Lemma 5** *For all $s \in [S]$:*

$$\psi_{s,T+1} \leq 2^s \eta \sqrt{\left(10 + \frac{15}{\gamma}\right)\tilde{d}\psi_{s,T+1}l_T}$$

The lemmas above have analysed the properties of BaseKernelUCB assuming independence. The effect of the SupKernelUCB construction is described in Lemma 14 and 15 by Auer (2002), which we restate for convenience using our notation. Lemma 6 shows that there the trials given to BaseKernelUCB are indeed independent:

**Lemma 6** *For each $s \in [S]$, each $t \in [T]$, and any fixed sequence of feature vectors $x_t$ with $t \in \Psi_t^{(s)}$, the corresponding rewards $r_t$ are independent random variables such that $\mathbb{E}(r_t) = \phi(x_t)^\intercal \theta^*$.*

Lemma 7 gives the properties of SupKernelUCB needed to provide the final regret bound, where the first item is a consequence of Lemma 1:

**Lemma 7** *With probability $1 - 2Ne^{-\eta^2/2}$, for any $t \in [T]$ and any $s \in [S]$, the following hold:*

- *$|\hat{\mu}_{a,t}^{(s)} - \phi(x_{a,t})^\intercal\theta^*| \leq (\eta(1+\|\theta^*\|) + \gamma^{1/2}\|\theta^*\|)\hat{\sigma}_{a,t}^{(s)}$, for all $a \in [N]$*

- *$a_t^* \in \hat{A}_s$, and*

- *$\mathbb{E}[r_{a_t^*,t}] - \mathbb{E}[r_t] \leq 2^{3-s}$ for any $a \in \hat{A}_s$.*

Now we find the final regret bound with a similar scheme as Auer (2002) using all the previous lemmas.

**Proof of Theorem 1.** First we upper bound the expected regret $\mathbb{E}[R(T)]$ with probability at least $1 - 2NTe^{-\eta^2/2}$ with:

$$\sum_{t \in [T]\backslash\Psi_0} \mathbb{E}(r_{a_t^*,t}) - \mathbb{E}(r_t) = \sum_{s=1}^{S} \sum_{t \in \Psi_{T+1}^{(s)}} \mathbb{E}(r_{a_t^*,t}) - \mathbb{E}(r_t)$$

$$\leq \sum_{s=1}^{S} 2^{3-s}\psi_{s,T+1} \leq \eta S \sqrt{\left(10 + \frac{15}{\gamma}\right)\tilde{d}\psi_{s,T+1}l_T} \quad (9)$$

and:

$$\sum_{t \in \Psi_0} \mathbb{E}(r_{a_t^*,t}) - \mathbb{E}(r_t) \leq 2\left(2 + \|\theta^*\| + \frac{\gamma^{\frac{1}{2}}\|\theta^*\|}{\eta}\right)\sqrt{T},$$

$$(10)$$

where $\Psi_0 := [T] \setminus \bigcup_{s \in [S]} \Psi_{T+1}^{(s)}$. In (9) we have used Lemma 7 for the first inequality and Lemmas 5 and 6 for the second inequality. In (10) we used Lemma 1 and the construction of the $\Psi_0$ set in SupKernelUCB. Now a standard application of the Azuma-Hoeffding inequality tells us that, with probability at least $1 - 2TN e^{-\eta^2/2}$:

$$|R(T) - \mathbb{E}(R(T|H_{T-1}))| \leq \sqrt{2T \ln(1/TN e^{-\eta^2/2})}. \quad (11)$$

Finally setting $S = \ln T$, $\eta = \sqrt{2 \ln 2TN/\delta}$, and bounding $\psi_{s,T+1}$ by $T$ we obtain from (9), (10), and (11):

$$R(T) = \mathbb{E}(R(T))) + [R(T) - \mathbb{E}(R(T|H_{T-1}))]$$
$$\leq 8 \sqrt{\left(10 + \frac{15}{\gamma}\right) l_T^3 \left(2 \ln \frac{2TN}{\delta}\right)} \sqrt{\tilde{d}T}$$
$$+ 2 \left(1 + \left(1 + \sqrt{\frac{\gamma}{2 \ln(2TN/\delta)}}\right) \|\theta^*\|\right) \sqrt{T}$$
$$+ \sqrt{2 \ln(1/\delta)} \sqrt{T}.$$

with probability $1 - (1 + \ln T)\delta$. The result follows by substituting $\delta/(1 + \ln T)$ for $\delta$. ∎

### 4.1 Relationship with GP-UCB

We now relate our analysis to that of GP-UCB in (Srinivas et al., 2010), and in particular to their Theorem 3, which treats the agnostic case. In this case, $\theta^*$ is not assumed to be sampled from a GP, but instead to have a bounded RKHS norm $\|\theta^*\|$. Under this assumption, the cumulative regret is bounded as:

$$O\left(\left(I(y_A; \theta^*) + \|\theta^*\|^2 \sqrt{I(y_A; \theta^*)}\right) \sqrt{T}\right), \quad (12)$$

where $I(y_T; \theta^*)$ is the mutual information between $\theta^*$ and the vector of (noisy) observations $y_T$. Both $I(y_T; \theta^*)$ in (12) and $\tilde{d}$ are data dependent quantities. We now relate them in order to compare the analyses. We have that:

$$I(y_T; f) = \ln|I + \sigma^{-2} K_T| = \sum_i \ln(1 + \sigma^{-2} \lambda_{i,T})$$
$$\geq \ln\left(1 + \sigma^{-2} \lambda_{\tilde{d}-1,T}\right)\left(\tilde{d} - 1 + \frac{\sum_{i > \tilde{d}-1} \lambda_{i,T}}{\lambda_{\tilde{d}-1,T}}\right)$$
$$\geq (\tilde{d} - 1)\ln\left(1 + \sigma^{-2} \lambda_{\tilde{d}-1,T}\right)\left[1 + \frac{\gamma \ln T}{\lambda_{\tilde{d}-1,T}}\right]$$
$$\geq (\tilde{d} - 1)\max_B \min\left\{\ln(1 + B)\gamma\sigma^{-2}\ln(T), \frac{\ln(1 + B)}{B}\right\}$$
$$\geq \Omega(\tilde{d}\ln\ln T)$$

In the second equality, we used the fact that the eigenvalues of $\Phi_T^\top \Phi_T$ are the same as the eigenvalues of

$\Phi_T \Phi_T^\top$. In the second inequality we used the definition of $\tilde{d}$. For the last inequality we considered the two cases when $\lambda_{\tilde{d}-1,T} \leq B\sigma^2$ and when $\lambda_{\tilde{d}-1,T} \geq B\sigma^2$ for some $B$.

This shows that $\tilde{d}$ is at least as good as $I(y_T; \theta^*)$, and comparing our Theorem 1 with (12), our regret bound only scales as $O(\sqrt{\tilde{d}})$, while the dependence of the regret bound (12) is linear in $I(y_T; \theta^*)$. In particular, this means that for the linear kernel we attain the lower bound for linear contextual bandits, (Chu et al., 2011), while GP-UCB is $\sqrt{d}$ away. This concerns only the agnostic case of GP-UCB, i.e. Theorem 3 in (Srinivas et al., 2010), which is the same setting as ours. When $\theta^*$ is sampled from a GP, their result for linear case also matches the lower bound.

Srinivas et al. (2010) also provide an upper bound on $I(y_T; \theta^*)$, denoted by $\gamma_T$, for certain kernels. As a consequence of the link between $I(y_T; \theta^*)$, $\gamma_T$ and $\tilde{d}$, we may also express our bounds in terms of $\gamma_T$. Moreover, in the agnostic case again, our bounds enjoy an improved dependence on this parameter: for example, for the widely used RBF kernel, our bound scales with $O(\ln T)^{d/2}$ in place of $O(\ln T)^d$.

Finally, when $\|\theta^*\|$ is unknown and we are unable to regularise appropriately, our regret bound only depends on $\|\theta^*\|$ linearly (Remark 3), while the dependence in (12) is quadratic.

## 5 Conclusion

We derive and analyse KernelUCB, an algorithm for contextual bandits, which is able to run with just a similarity function instead of context features. We give a finite-time theoretical analysis that proves the cumulative regret scales as $\tilde{O}(\sqrt{T\tilde{d}})$ where $\tilde{d}$ is the effective dimension of the data in the feature space.

As a special case of our algorithm and its analysis, we recover the known upper bound for LinUCB that matches the lower bound for the linear problem. In the case when we know an upper bound on the model noise, then setting our regulariser to that value recovers the GP-UCB algorithm. Moreover, we provide an improved analysis for the agnostic case, when the reward function is not necessarily sampled from a GP prior. Finally, our analysis shows the dependence of the regulariser on the RKHS norm of reward function.

## 6 Acknowledgements

This research was funded by European Community's Seventh Framework Programme (FP7/2007-2013) under grant agreement n$^o$ 270327 (project CompLACS).